\documentclass[sigconf]{acmart}
\AtBeginDocument{%
  \providecommand\BibTeX{{%
    \normalfont B\kern-0.5em{\scshape i\kern-0.25em b}\kern-0.8em\TeX}}}

\copyrightyear{2023}
\acmYear{2023}
\setcopyright{rightsretained} 
\acmConference[GoodIT '23]{ACM International Conference on Information Technology for Social Good}{September 6--8, 2023}{Lisbon, Portugal}
\acmBooktitle{ACM International Conference on Information Technology for Social Good (GoodIT '23), September 6--8, 2023, Lisbon, Portugal}
\acmDOI{10.1145/3582515.3609559}
\acmISBN{979-8-4007-0116-0/23/09}

\usepackage{float}

\begin{document}

\title[CliniDigest: LLM-Based Large-Scale Summarization of Clinical Trials]{CliniDigest: A Case Study in Large Language Model Based Large-Scale Summarization of Clinical Trial Descriptions}

\author{Renee D. White}
\email{reneedw@cs.stanford.edu}
\orcid{0009-0009-1225-1580}
\affiliation{%
  \institution{Stanford University}
  \streetaddress{450 Serra Mall}
  \city{Stanford}
  \state{California}
  \country{USA}
  \postcode{94305}
}
\authornote{Co-first author}

\author{Tristan Peng}
\email{tristan@cs.stanford.edu}
\orcid{0000-0001-7907-574X}
\affiliation{%
  \institution{Stanford University}
  \streetaddress{450 Serra Mall}
  \city{Stanford}
  \state{California}
  \country{USA}
  \postcode{94305}
}
\authornotemark[1]

\author{Pann Sripitak}
\email{pannsr@cs.stanford.edu}
\orcid{0009-0003-1352-0494}
\affiliation{%
  \institution{Stanford University}
  \streetaddress{450 Serra Mall}
  \city{Stanford}
  \state{California}
  \country{USA}
  \postcode{94305}
}

\author{Alexander Rosenberg Johansen}
\email{arjo@stanford.edu}
\orcid{0000-0002-4993-7916}
\affiliation{%
  \institution{Stanford University}
  \streetaddress{450 Serra Mall}
  \city{Stanford}
  \state{California}
  \country{USA}
  \postcode{94305}
}
\authornote{Co-corresponding author}

\author{Michael Snyder}
\email{mpsnyder@stanford.edu}
\orcid{0000-0003-0784-7987}
\affiliation{%
  \institution{Stanford University}
  \streetaddress{450 Serra Mall}
  \city{Stanford}
  \state{California}
  \country{USA}
  \postcode{94305}
}
\authornotemark[2]

\renewcommand{\shortauthors}{White, Peng, et al.}

\begin{abstract}
    A clinical trial is a study that evaluates new biomedical interventions.
    To design new trials, researchers draw inspiration from those current and completed.
    In 2022, there were on average more than 100 clinical trials submitted to ClinicalTrials.gov every day, with each trial having a mean of approximately 1500 words \cite{clinicaltrials.gov}. This makes it nearly impossible to keep up to date.
    To mitigate this issue, we have created a batch clinical trial summarizer called CliniDigest using GPT-3.5.
    CliniDigest is, to our knowledge, the first tool able to provide real-time, truthful, and comprehensive summaries of clinical trials.
    CliniDigest can reduce up to 85 clinical trial descriptions (approximately 10,500 words) into a concise 200-word summary with references and limited hallucinations.
    We have tested CliniDigest on its ability to summarize 457 trials divided across 27 medical subdomains. For each field, CliniDigest generates summaries of $\mu=153,\ \sigma=69 $ words, each of which utilizes $\mu=54\%,\ \sigma=30\% $ of the sources.
    A more comprehensive evaluation is planned and outlined in this paper.
\end{abstract}



\begin{teaserfigure}
    \centering
  \includegraphics[width=0.8\textwidth]{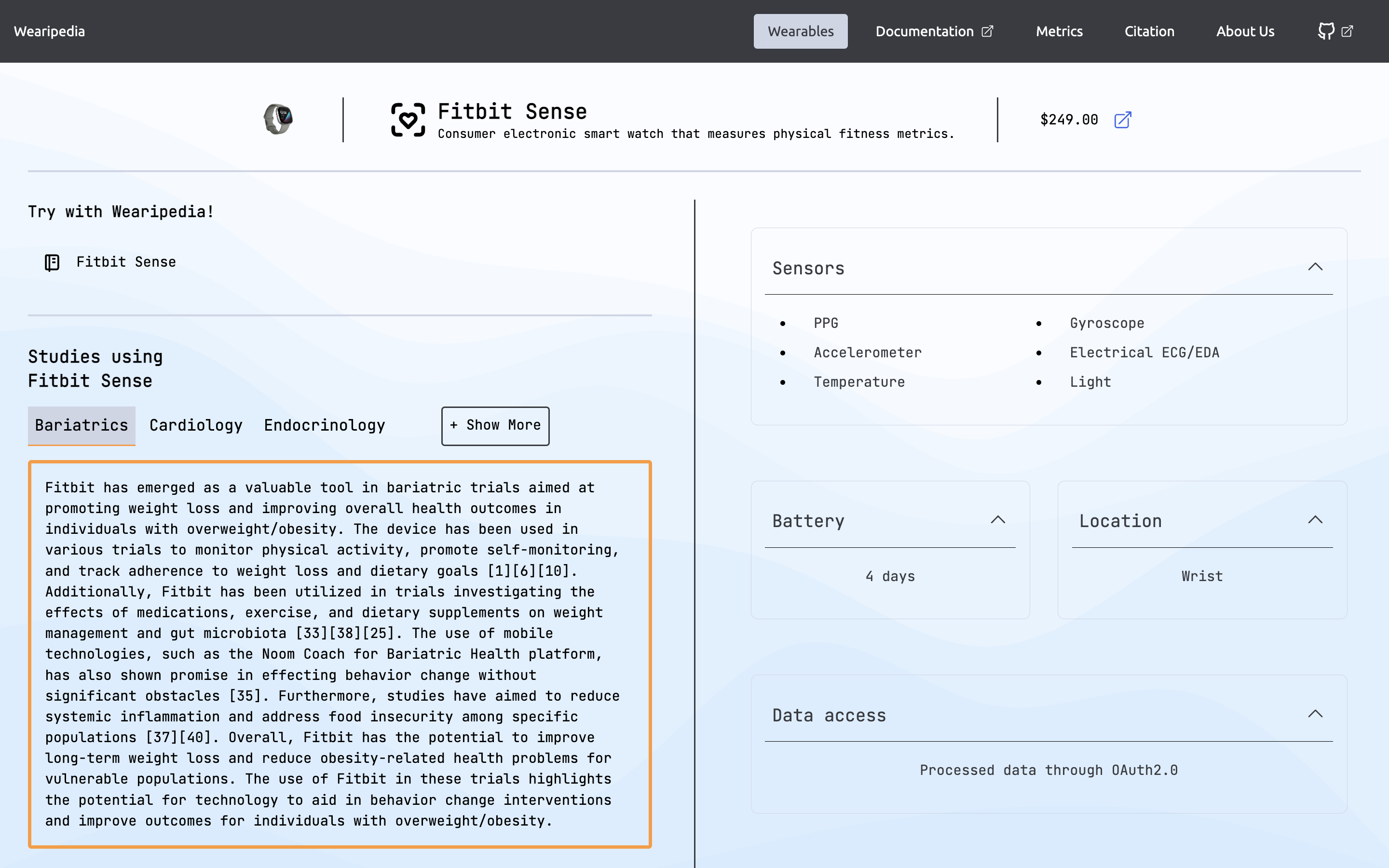}
  \caption{CliniDigest integrated into Wearipedia. The text in the orange box on the left of the page is automatically synthesized by CliniDigest using data from ClinicalTrials.gov. For this particular example, the text summarizes the Fitbit Sense wearable's use in clinical research in bariatrics.}
  \Description{Wearipedia home page.}
  \label{fig:teaser}
\end{teaserfigure}

\received{1 June 2023}
\maketitle

\section{Introduction}
\label{sec:intro}
Clinical trials are the de facto method for developing new treatments and innovations within most medical fields.
On average, clinical trials take 21 months from being posted on MEDLINE and ClinicalTrials.gov to publication \cite{Ross_Mocanu_Lampropulos_Tse_Zarin_Krumholz_2013}.
Moreover, only about 46\% of completed studies publish articles. Reasons for not publishing vary from publication bias, commercial interest, to lack of significance \cite{Ross_Mulvey_Hines_Nissen_Krumholz_2009}.
However, ClinicalTrials.gov maintains a list of all trials published, unpublished, and in progress; clinical research coordinators have to register their studies with ClinicalTrials.gov when they start recruiting participants and keep the entry updated until the trial's completion or withdrawal. 
This makes ClinicalTrials.gov a valuable resource to get up-to-date information on current clinical trials. 
However, the increase in clinical trial submissions creates a need for search tools to compile the vast amount of information.
We present CliniDigest, a prompt engineering technique for GPT-3.5 that condenses up to 85 clinical trials into a 200-word summary with references.
The 457 clinical trials used as our testing dataset for CliniDigest are scraped from ClinicalTrials.gov and are annotated into 14 medical fields and two subdomains: completed and new.
Tables \ref{tab:prompt} and \ref{tab:combo prompt} show example prompts and figure \ref{fig:teaser} and table \ref{tab:combo results} show summaries of CliniDigest. 
Since GPT-3.5 is limited to 4,096 tokens, we use multiple levels of summarization and prompting to get truthful and concise summaries, explained in more detail in section \ref{sec:methodology}.
Upon completion, CliniDigest, the subject of this paper, will be available on Wearipedia (\url{https://www.wearipedia.com}), which is an open-sourced, encyclopedic directory of biomedical fitness sensors and trials using them.
Currently, Wearipedia only includes a set of links to raw clinical trials scraped for each device using an advanced text-search algorithm \cite{Marra_Chen_Coravos_Stern_2020}.

Beyond providing in-depth information about wearables, Wearipedia also provides a \href{https://pypi.org/project/wearipedia/}{Python package} for a uniformed data extraction process across all supported wearables.\footnote{Code for the library can be found here: \url{https://github.com/Stanford-Health/wearipedia}.} Accompanying the Python libraries are premade Jupyter Notebooks that fully demonstrate the capabilities of Wearipedia's data extraction pipeline.
Additionally, they provide an explanatory boilerplate for users more unfamiliar with programming or those who want a quick start to easily extract data from their wearables.
Given the pedigree of Wearipedia, accessibility and comprehensiveness of wearable data and information are the core tenets of the project. Thus, it is crucial that the information that Wearipedia presents on each wearable's device page is as accurate, complete, and digestible as possible.
The tensions between Wearipedia's goals as a whole with the enormity of data and amount of wearables suggest that the conventional solution of manually crafting thorough abstracts for the purpose of Wearipedia's goals does not cater well to the overall scale and progression of the project, motivating the development of CliniDigest.
CliniDigest will be used by the clinical team, which often includes physicians, researchers, clinical research coordinators worldwide. Since these team members come from a wide spectrum of academic backgrounds and need the information in a timely manner, we have the following goals: accessibility, efficiency, accuracy, comprehensiveness, and timeliness.

First, the summaries should be accessible by any member of the clinical trial community, which covers a diverse group of individuals. Accessibility is determined by readability and empirical evaluation by clinical team members. Second, as a part of Wearipedia, CliniDigest should expedite the process of choosing the right wearable for a clinical question. Furthermore, a summary should not misrepresent clinical trials and although a summary may not be able to include all trials, it should cover as broadly as possible. Moreover, as a major benefit of ClinicalTrials.gov is access to trials before official publication, CliniDigest should reflect updates and changes when they happen.

CliniDigest represents an example of integrating GPT-3.5 into a text-centric database. It is able to solve text summarization in a fashion that GPT-3.5, information retrieval, or text summarization algorithm wouldn't be able to do on their own.

We have done preliminary evaluations on the length, reference usage, and readability of the summaries.
We have also planned an extensive evaluation that can empirically test the efficacy of CliniDigest's large-scale compression. This would test the ability of clinical research coordinators to understand the breadth and depth of their investigative field.
A successful evaluation would entail the integration of CliniDigest into the pipeline that transforms raw, web-scraped clinical trial data from various sources into a refined summary displayed in an intuitive user interface on Wearipedia's website.

\begin{figure*}[!ht]
    \includegraphics[width=0.6\textwidth]{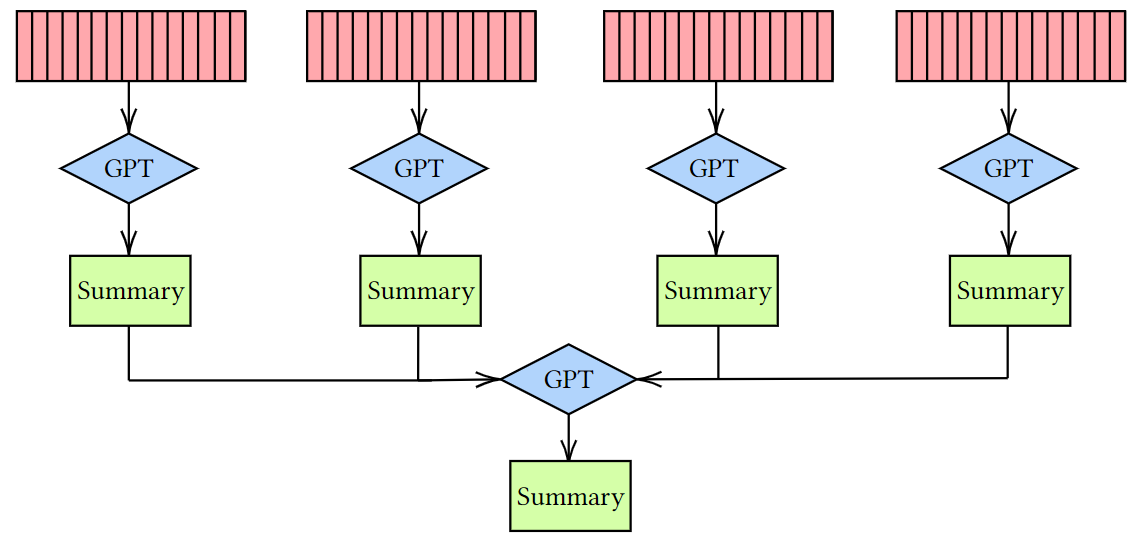}\centering
    \caption{Batch summarization method: we circumvent the token limit through a series of cascading inputs through GPT-3.5. First, we split the raw data into batches of 15 studies each. Then, we prepend the intermediate prompt shown in table \ref{tab:prompt} as a prompt to GPT-3.5. Thus, each batch will produce one summary. After each batch has been processed, we combine the intermediary summaries through the prompt shown in \ref{tab:combo prompt} to further condense the summary into a concise description that encompasses the extent of our data.}
    \Description{Batch Summarization Diagram}
  \label{fig:batch}
\end{figure*} %

\section{Related Work}
\label{sec:related work}
Text summarization of medical literature is an active field of research. Due to the exponential rise in registered clinical trials, research has shifted from single-document summarization to multi-document summarization, with techniques trending to those incorporating machine learning \cite{See_Liu_Manning_2017, Mishra_Bian_Fiszman_Weir_Jonnalagadda_Mostafa_Del_Fiol_2014}. There are four main types of summarization, grouped on two axes: \textit{extractive} (summarization through extracting source material) versus \textit{abstractive} (summarization through paraphrasing the main idea), and \textit{indicative} (providing a general overview of the topic) versus \textit{informative} (providing in-depth information) \cite{Mishra_Bian_Fiszman_Weir_Jonnalagadda_Mostafa_Del_Fiol_2014}. For CliniDigest, we strive to create \textit{indicative abstractive} summaries, differing from the extractive paradigm that has already shown to be effective in summarization of clinical trials \cite{Gulden_Kirchner_Schuttler_Hinderer_Kampf_Prokosch_Toddenroth_2019}.

Recent advances in summarization have been led by the emergence of general purpose large language models \cite{Kieuvongngam_Tan_Niu_2020, Afantenos_Karkaletsis_Stamatopoulos_2005, Mishra_Bian_Fiszman_Weir_Jonnalagadda_Mostafa_Del_Fiol_2014}. Through prompt engineering, the style of summarization can be adapted to many types of writing and summary styles.
Prompt engineering allows users to efficaciously converse with large language models: through meticulous design of these instructions, users can ensure the quality and form of a response. This rose out of recent developments in natural language processing that found it more cost-effective and efficient to train generalizable prompt-based models based on evaluating text probability, one that GPT-3.5 falls under \cite{Liu_Yuan_Fu_Jiang_Hayashi_Neubig_2023, Lester_Al-Rfou_Constant_2021, Brown_Mann_Ryder_Subbiah_Kaplan_Dhariwal_Neelakantan_Shyam_Sastry_Askell_etal._2020}. Although the field is still relatively new, much scholarly work has already been done in the standardization and categorization of effective prompt engineering paradigms \cite{White_Fu_Hays_Sandborn_Olea_Gilbert_Elnashar_Spencer-Smith_Schmidt_2023}. Other scholarly endeavors have examined the ability of large language models to enhance efficiency and performance in various information-transfer based applications, such as content creation via refining social media and crowdsourcing pitches, assisting with the writing of important sections in radiology reports, generation of high-quality and empathetic answers to patient questions, and low variance summarization and zero-shot de-identification of patient records \cite{Short_Short_2023, Ma_Wu_Wang_Xu_Wei_Liu_Guo_Cai_Zhang_Zhang_etal._2023, Ayers_Poliak_Dredze_Leas_Zhu_Kelley_Faix_Goodman_Longhurst_Hogarth_etal._2023, Wang_Shi_Yu_Wu_Ma_Dai_Yang_Kang_Wu_Hu_etal._2023, Chuang_Tang_Jiang_Hu_2023, Liu_Yu_Zhang_Wu_Cao_Dai_Zhao_Liu_Shen_Li_etal._2023}.

Despite the promises of using prompt engineering, there are many pitfalls to watch out for. Coined by research in neural machine translation, but similarly applied to prompt-based large language models, the GPT-4 technical paper describes hallucinations as the tendency to ``\thinspace`produce content that is nonsensical or untruthful in relation to certain sources'\thinspace'' \cite{Agarwal_Wong-Fannjiang_Sussillo_Lee_Firat_2018, OpenAI_2023}. This type of output is categorically antithetical to the goals of the project. Although formally defined in the GPT-4 technical paper, GPT-3 (and subsequently GPT-3.5) also faces similar limitations hallucinating in both open- and closed-domain contexts by providing informative yet misleading responses \cite{Maynez_Narayan_Bohnet_McDonald_2020, OpenAI_2023, Lin_Hilton_Evans_2022}. Nonetheless, GPT-3 has shown strong performance in $n$-shot prompting tasks, making it a strong suitor, and the model of choice, for this project \cite{Brown_Mann_Ryder_Subbiah_Kaplan_Dhariwal_Neelakantan_Shyam_Sastry_Askell_etal._2020}. Beyond hallucinations, it is also a uniquely challenging problem for researchers to design satisfactory human-AI interactions due to two main factors: uncharted capabilities of AI and its output complexity \cite{Yang_Steinfeld_Rose_Zimmerman_2020}.

For medical literacy, the SMOG test has shown to be superior on consistency, accuracy, and speed for medical literature in comparison with other standard metrics such as Flesch-Kincaid, FOG, Flesch Reading Ease, Fry, and Dale-Chall \cite{McLaughlin_1969, Wang_Miller_Schmitt_Wen_2013}. In the larger planned evaluation, we also plan to use the ROUGE-L F1 metric to evaluate the quality of the synthesized summaries \cite{Lin_2004}.


\setlength\intextsep{2mm}{
\begin{table*}[!ht]
    \centering
    \scalebox{0.8}{%
        \begin{tabular}{|p{1.25\textwidth}|}
            \hline
            \textbf{Prompt} - single summary\\
            \hline
            Your task is to extract relevant information from 15 trials delimited in the triple backticks labeled from 1 to 15 to construct an argument about the purpose of Fitbit in general physiology trials. Each trial contains a title and description. Your reader will be clinical research coordinators.
            \newline\newline
            Write a 200 word thesis with references to the trials in the following format: [1].
            \newline\newline
            Trials: \`{}\`{}\`{}[Redacted]\`{}\`{}\`{}
            \\
            \hline
        \end{tabular}
    }
    \caption{Example prompt for a single summary for the uses of Fitbits in general physiology.}
    \Description{
    Prompt - single summary: Your task is to extract relevant information from 15 trials delimited in the triple backticks labeled from 1 to 15 to construct an argument about the purpose of Fitbit in general physiology trials. Each trial contains a title and description. Your reader will be clinical research coordinators.
            Write a 200 word thesis with references to the trials in the following format: [1].
            Trials: \`{}\`{}\`{}[Redacted]\`{}\`{}\`{}}
    \label{tab:prompt}
\end{table*}
}

\setlength\intextsep{0mm}{
\begin{table*}[!ht]
    \centering
    \scalebox{0.8}{%
        \begin{tabular}{|p{1.25\textwidth}|}
            \hline
            \textbf{Prompt} - multi summary\\
            \hline
            Your task is to extract relevant information from the provided text and references to
            construct a cumulative argument about the purpose of Fitbit in general physiology trials.
            Each paragraph includes references to clinical trials in the following format: [1].
            Weigh each paragraph according to its word count, weighing
            longer paragraphs more than shorter ones. Your reader will be clinical research coordinators. \\ 
            Summary: \`{}\`{}\`{}[Redacted] \`{}\`{}\`{}\\
            References: \`{}\`{}\`{}[Redacted] \`{}\`{}\`{}\\
            Write a 150-250-word thesis with references to the trials in the following format: [1]. \\
            \hline
        \end{tabular}
    }
    \caption{Example concatenated prompt for the uses of Fitbits in general physiology.}
    \Description{\textbf{Prompt} - multi summary: Your task is to extract relevant information from the provided text and references to
            construct a cumulative argument about the purpose of Fitbit in general physiology trials.
            Each paragraph includes references to clinical trials in the following format: [1].
            Weigh each paragraph according to its word count, weighing
            longer paragraphs more than shorter ones. Your reader will be clinical research coordinators.
            Summary: \`{}\`{}\`{}[Redacted] \`{}\`{}\`{}\\
            References: \`{}\`{}\`{}[Redacted] \`{}\`{}\`{}\\
            Write a 150-250-word thesis with references to the trials in the following format: [1].}
    \label{tab:combo prompt}
\end{table*}
}\setlength\intextsep{0mm}{

\section{Methodology}
\label{sec:methodology}

In order to get the summaries we had a multi-step pipeline. First, we extracted all studies related to Fitbit from ClinicalTrials.gov \cite{Marra_Chen_Coravos_Stern_2020}. Of these, we reviewed and annotated 457 trials into its primary medical field, of which we include: somnology, gynecology, obstetrics, cardiology, general physiology, endocrinology, bariatrics, psychiatry, oncology, gastroenterology, pulmonology, chronic pain/diseases, nephrology, and other.
Moreover, we separated the studies into completed (within the past 5 years) and new (within the past two years). We also excluded trials that were withdrawn and trials with enrollment numbers less than 50. 
This reduction led us to between 1-84 clinical studies for 27 medical field combinations. Nephrology, as the only one, did not have any studies for completed trials. This fine-grained sorting of clinical trials was then fed into GPT-3.5 to make 200-word summaries.

Since GPT-3.5 Turbo has a maximum token limit of 4,096, even just clinical trials associated with one wearable in one medical field exceeds this limit. For example, a hand-picked subset of 25 oncology trials using the Fitbit physical fitness monitor well exceeds the restrictive 4,096 token limit; aggregating them into a compact list of just trial names and brief descriptions, the example subset works out to be nearly 4,900 tokens. Thus, one of the biggest obstacles, and biggest design challenge we faced, was to design a way to circumvent this limit through clever prompting that would yield a similar result without pushing against the limit. To this end, we split the studies into batches of 15 clinical trials, which were recursively summarized until it was condensed into one succinct summary. In the case that a batch had less than 15 clinical trials, we adjusted the prompt's word count to be proportional to the number of clinical trials inputted (13 words per trial).
Figure \ref{fig:batch} shows an example workflow for a list of trials subdivided into four batches, which are iterated upon with GPT-3.5 until it reaches one condensed summary.
{

For the prompt design, we first prioritized a design that promotes readability. To do so, we constructed our prompt in a way that instructs GPT-3.5 Turbo to use language accessible to clinical research coordinators. The inputted trials, as referred to as Trials: \`{}\`{}\`{}[Redacted]\`{}\`{}\`{} in table \ref{tab:prompt}, each consists of the clinical trial's title and its brief description listed on ClinicalTrials.gov. As the brief descriptions usually contain predominantly medical jargon, summarization without this language simplification specification is not sufficient for our goal of making clinical trial data accessible to clinical research coordinators. 

Second, we wanted to develop a prompt that is able to synthesize broad concepts relating to all trials. This marked a major challenge because lackadaisical prompts often caused GPT-3.5 to simply summarize every study individually. To mitigate this, we experimented with a variety of prompts that prioritized generalization through keywords such as ``summary,'' ``essay,'' and ``argument.'' We found that ``thesis'' and ''argument' work the best, which can be seen in table \ref{tab:prompt}.

Another design challenge we needed to address was circumventing a crucial limitation of such models: hallucinations. As discussed in section \ref{sec:related work}, hallucinations are the bane to effective and accurate summarization of clinical trials, as it would not only spread misinformation, but it would also be difficult to comprehensively fact-check every source for accuracy. To address this, we prompted GPT-3.5 to use references whenever possible, as can be seen in tables \ref{tab:prompt} and \ref{tab:combo prompt}.
We found that tuning the temperature, which controls creativity, to 0 and prompt engineering for explainability reduced hallucinations. While not a perfect solution, it still aims to reduce the number of hallucinations that occur in these closed-context circumstances without large modifications to the overall model. Although large language models are not designed to be factual, by asking for references, GPT is required to utilize the inputted clinical trials.

\section{Results}
\label{sec:results}
In table \ref{tab:combo results}, we see an example output from GPT-3.5 that summarizes 39 completed clinical trials focusing on general physiology. The output includes 11 references in 201 words. An informal analysis shows that this output is promising, as it specifically references general physiology trials, generalizes to include multiple trials, and provides useful information for a clinical research coordinator. Moreover, it includes in-text references with relatively uniformly spread citations in addition to explaining key characteristics of specific trials without naming them (which would take up words in our limited 150-250 output and is unnecessary because of the inclusion of references). 

\setlength\intextsep{2mm}{
\begin{table*}[!ht]
    \centering
    \scalebox{0.8}{%
        \begin{tabular}{|p{1.25\textwidth}|}
            \hline
            \textbf{Result}\\
            \hline
            Fitbit has emerged as a valuable tool in General Physiology trials aimed at promoting physical activity and improving health outcomes. Fitbit is a wearable activity tracker that allows researchers to monitor physical activity levels and track progress towards set goals. It has been used in a variety of trials, including those focused on cancer survivors [1], adolescents [6], and older adults [10]. Fitbit has also been utilized in trials that aim to promote physical activity in specific populations, such as Veterans with lower-limb amputation [5] and South Asian immigrant women and girls [14]. In addition to promoting physical activity, Fitbit has been used in trials focused on improving muscle function [8], treating depressive symptoms [2], and managing chronic pain [15]. Fitbit has been used in conjunction with other interventions, such as resistance training and virtual reality therapy, to improve health outcomes. Furthermore, Fitbit has been utilized to support lifestyle physical activity interventions and improve health outcomes in trials [31], [32], and [33]. These trials demonstrate the potential of Fitbit to support physical activity interventions and improve health outcomes. Overall, Fitbit has become a valuable tool in General Physiology trials aimed at promoting sustained exercise participation and improving physical and psychosocial well-being. \\
            \hline
        \end{tabular}
    }
    \caption{Final results: this is a summary on Fitbit's use in general physiology trials.}
    \Description{Result: Fitbit has emerged as a valuable tool in General Physiology trials aimed at promoting physical activity and improving health outcomes. Fitbit is a wearable activity tracker that allows researchers to monitor physical activity levels and track progress towards set goals. It has been used in a variety of trials, including those focused on cancer survivors [1], adolescents [6], and older adults [10]. Fitbit has also been utilized in trials that aim to promote physical activity in specific populations, such as Veterans with lower-limb amputation [5] and South Asian immigrant women and girls [14]. In addition to promoting physical activity, Fitbit has been used in trials focused on improving muscle function [8], treating depressive symptoms [2], and managing chronic pain [15]. Fitbit has been used in conjunction with other interventions, such as resistance training and virtual reality therapy, to improve health outcomes. Furthermore, Fitbit has been utilized to support lifestyle physical activity interventions and improve health outcomes in trials [31], [32], and [33]. These trials demonstrate the potential of Fitbit to support physical activity interventions and improve health outcomes. Overall, Fitbit has become a valuable tool in General Physiology trials aimed at promoting sustained exercise participation and improving physical and psychosocial well-being.}
    \label{tab:combo results}
\end{table*}
}

Across all 14 fields, CliniDigest generates summaries with $\mu=153,\ \sigma=69 $ length and each summary utilizes $\mu=54\%,\ \sigma=30\% $ of the sources. In the 17 summaries that required the second combination step, the final summaries have $\mu=192,\ \sigma=27$ words with 100\% of them following between the requested 150-250 word range. The distribution can be seen in figure \ref{fig:referencedist}.

\begin{figure*}[!ht]
    \includegraphics[width=0.6\textwidth]{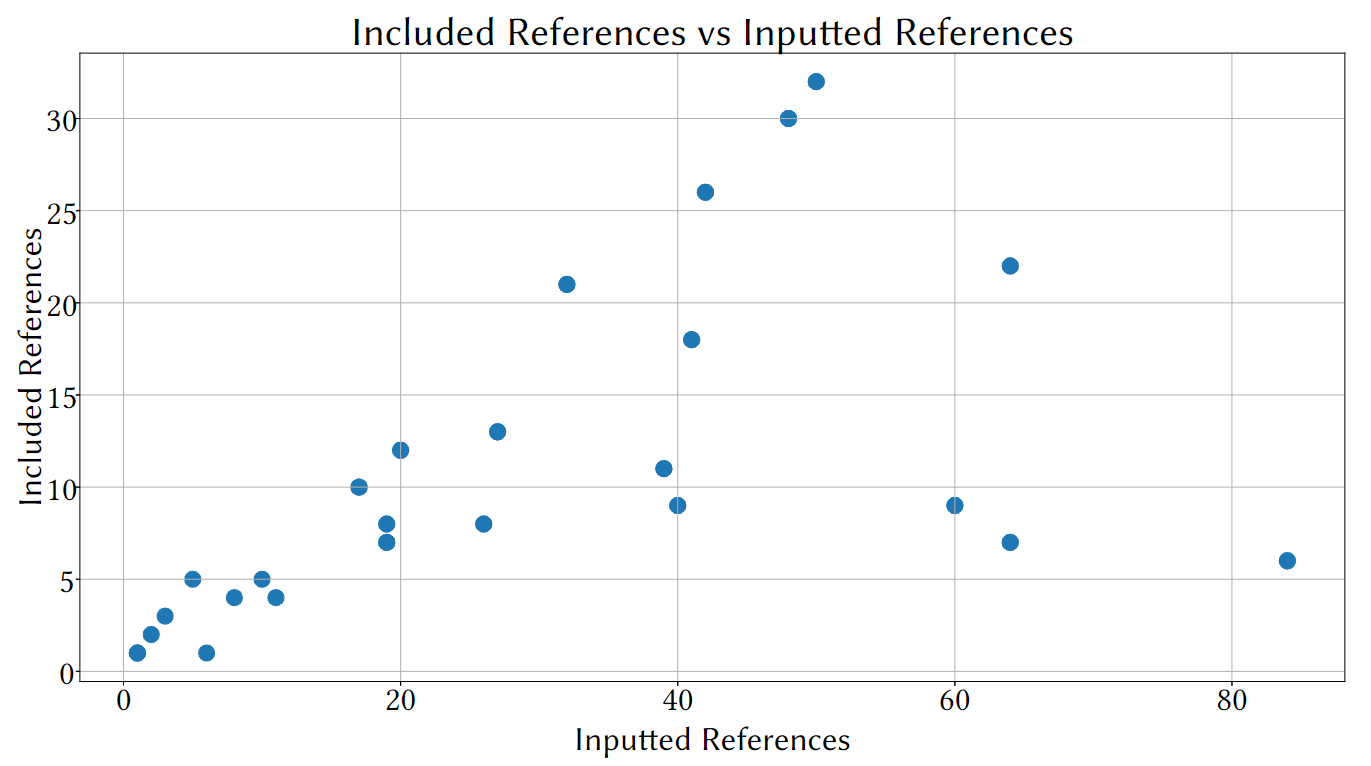}\centering
    \caption{Scatter plot of the number of references included in the final output summaries compared to the number of inputted trial descriptions. There is a general linear trend up to 50 trials with slope $m=0.5140$ and correlation coefficient $r^2=0.7547$. After the inflection point of around 50 trials, there are diminishing returns to the number of sources that CliniDigest includes in its summaries due to the inherent difficulty of compressing a large number of trials within a small word limit.}
    \Description{Plot of Reference Inclusion}
  \label{fig:referencedist}
\end{figure*} %

\subsection{Readability}
\label{par:readability}
In order to ensure effective communication, health information materials need to be written in an understandable manner. To evaluate the readability of the GPT-3.5 outputs, we utilized the SMOG readability formula \cite{McLaughlin_1969, Wang_Miller_Schmitt_Wen_2013}. As a quantitative test of accessibility, we conducted a two-sample T-test of means to compare two corpora: the original study titles and descriptions, and the summaries synthesized by CliniDigest. Using the SMOG formula, we calculated $\bar{x}_1=19.32$ and $s_1=1.220$ for raw data from ClinicalTrials.gov, and $\bar{x}_2=18.49$ and $s_2=2.148$ for CliniDigest summaries. In all, $p=0.0929$ so we find that CliniDigest summaries are not significantly different in readability from the original trial descriptions.

This result still demonstrates some important points about CliniDigest's summaries. First, it shows that CliniDigest's summaries maintain the level of diction as its constituent raw descriptions: despite the condensation of upwards of 80 studies into a neat 200-word output, CliniDigest did not saturate its summaries with polysyllabic medical jargon to help compress information. Since researchers are expected to be able to read at this level, CliniDigest, according to the SMOG metrics, should be something they would be familiar with reading about.

While the SMOG measure provides quantitative evidence of CliniDigest's efficacy, readability formulae are not authoritative, but rather act as guidelines that serve as an addendum to an actual evaluation such as one planned in section \ref{sssec:evaluation plan} \cite{Meade_Smith_1991}. As for SMOG specifically, there are a variety of factors of CliniDigest that may inadvertently raise its SMOG score, causing an artificially inflated score. First, the conciseness of CliniDigest's summaries increases the density of polysyllabic words per sentence, the crucial metric that SMOG operates on for readability. Here we define polysyllabic as words with three or more syllables for concision and conformity of SMOG's definitions. This is compounded by the fact that with regards to clinical trials involving wearables, there are an inescapable amount of polysyllabic words that should be incorporated. Since many of these polysyllabic words are at the core of a clinical trial's purpose and an integral part of a clinical researcher's lexicon, it would be unreasonable to expect that these words be substituted with other monosyllabically-composed phrases. Thus, these two important factors---compression and the inclusion of critical polysyllabic words and phrases---combine to contribute to a higher SMOG score.
\subsection{Systematic Evaluation Plan}
\label{sssec:evaluation plan}

We expect that the majority of Wearipedia's users are researchers, clinicians, clinical research coordinators, or other health professionals that seek to use wearables as a part of studies, trials, or patient monitoring. Thus, our evaluation will be focused on catering towards this population. First, we plan to recruit researchers who are conducting or planning to conduct research involving wearables according to best practices \cite{Caine_2016}. This could be easily achieved through Wearipedia's current userbase, but we believe that it is also important to include in our participants people that are not familiar with Wearipedia and its facilities. After the recruiting process, we will survey the participants on their experience as clinical research coordinators. Among the various questions will include their highest academic degree, time working as a clinical research coordinator, as well as experience with using wearables both in daily life and in clinical trials.

Second, each participant will be asked to learn about two distinct wearables that they are not as familiar with, in two different medical fields that they are not familiar with. We will label each wearable-field pair as a test for the clarity of this paper. By working outside their area of expertise, this will eliminate confounding variables corresponding to prior knowledge as well as provide a lower bound for the efficacy of the paradigm in concisely educating the user about the status quo of various wearables in different areas of study. Among these two tests, we plan to randomly assign one as a control, and another as an experimental test. The control test will be the second method of presentation as introduced in section \ref{sec:intro}: links on the sidebar of the device page that collect the studies in a list. On the other hand, the experimental test is solely the generated output through GPT-3.5 Turbo without any modifications or external links that are provided, except for the list of references that is provided at the end of each output containing the name of each trial summarized. The participants will then be asked to read both tests one at a time in a random permutation. This is done in order to eliminate any confounding variables regarding the order that the control and experiment are presented in. After the participants feel like they understand the information given to them in each test, they will then be asked to write a brief summary of what they have learned. We hope that through this writing exercise, participants can reflect and demonstrate their understanding of the material in both breadth and depth. This portion will be timed, but participants are encouraged to use as much time as they feel they need to write a tenable summary. Furthermore, we will use these participant-written summaries as part of a more quantitative measure by using them as part of computing the standard ROUGE metric (specifically the ROUGE-L F1 score) that compares human-written responses with those synthesized by CliniDigest \cite{Lin_2004, Gulden_Kirchner_Schuttler_Hinderer_Kampf_Prokosch_Toddenroth_2019}. By using participants' responses, this will eliminate confounding variables that arise from in-team human-written summaries that conform with summaries written by CliniDigest that artificially increase the ROUGE score.
After the participants finish both tests, the evaluation concludes. At this point, we plan to conduct a post-evaluation questionnaire as a debrief that targets the more qualitative experience of doing both tests. During this phase, we plan to ask about the accessibility of both methods beyond the metrics used above in section \ref{par:readability} as well as the physical ease of access of using both, the perceived challenges and advantages of using each method in terms of breadth and depth, as well as enjoyment of using each. This qualitative data, along with the time spent on each test and the manual assessment of the participants' responses to the evaluation prompt, we hope will give us a comprehensive picture of the true value of the summaries in advancing the goals of accessibility and efficiency.
\section{Future Work and Conclusion}
While this paper has shown the potential for real-time synthesis of clinical trials in a digestible and accurate medium, it must be corroborated by empirical findings through a rigorous evaluation. Section \ref{sssec:evaluation plan} outlines the evaluation plan for Wearipedia's trial summarizer - CliniDigest. Beyond a comprehensive evaluation, we will continue to monitor GPT-3.5 Turbo's outputs for accuracy and concision with all wearables and medical fields as well as refine the prompts for better summaries.

Another direction we are looking to explore regarding the usage of GPT's text completion paradigm rather than that of text summarization. As speculation, there could be multiple benefits to such an approach. First, a human-dictated structure could be better at prompting GPT in its summary. Moreover, this more controlled environment would produce less variability between summaries of different wearables and fields. This homogeneity would increase the overall cohesion of the project, as users would be able to develop a clearer mental model of the organizational structure of the abstracts.

Finally, while it is essential to strike a balance between readability and preserving the accuracy of the final results, efforts should be directed towards finding alternative ways to improve the overall readability without compromising the validity of the content.

In this paper, we have presented a case study in large-scale accurate and comprehensive summarization of clinical trials through the development of CliniDigest. From the initial promising results, we will continue to develop this project from a case study to actual implementation as a crucial centerpiece of Wearipedia. With the rapid burgeoning usage of wearables in clinical trials as well as the exponentially increasing trend of the number of clinical trials being proposed and conducted, these large language models prove to be an effective and timely tool for presenting the vast amount of knowledge in a dense and easily-readable format. As a crucial component of the Wearipedia project, these summaries allow Wearipedia to progress in line with its goals of accessibility and comprehensiveness, and gives users more context to work with when using Wearipedia to evaluate and test wearable devices.


\bibliographystyle{ACM-Reference-Format}
\bibliography{sample-base}

\end{document}